\documentclass[format=sigconf]{acmart}

\usepackage{booktabs} 

\usepackage{graphicx}
\usepackage{subfigure}
\usepackage{tikz}
\usepackage{amsmath}
\usepackage{amssymb}
\usepackage{multirow}
\usepackage{bm}
\usepackage{makecell}

\copyrightyear{2018} 
\acmYear{2018} 
\setcopyright{acmcopyright}
\acmConference[KDD '18]{The 24th ACM SIGKDD International Conference on Knowledge Discovery \& Data Mining}{August 19--23, 2018}{London, United Kingdom}
\acmBooktitle{KDD '18: The 24th ACM SIGKDD International Conference on Knowledge Discovery \& Data Mining, August 19--23, 2018, London, United Kingdom}
\acmPrice{15.00}
\acmDOI{10.1145/3219819.3219962}
\acmISBN{978-1-4503-5552-0/18/08}

\title[Transcribing Content from Structural Images with Spotlight Mechanism]{Transcribing Content from Structural Images with\\Spotlight Mechanism}

 \author{Yu Yin, Zhenya Huang}
 \affiliation{
   \institution{Anhui Province Key Laboratory of Big Data Analysis and Application, University of Science and Technology of China}
 }
 \email{{yxonic,huangzhy}@mail.ustc.edu.cn}

 \author{Enhong Chen}
 \authornote{The corresponding author.}
 \affiliation{
 	\institution{Anhui Province Key Laboratory of Big Data Analysis and Application, University of Science and Technology of China}
 }
 \email{cheneh@ustc.edu.cn}

 \author{Qi Liu}
 \affiliation{
 	\institution{Anhui Province Key Laboratory of Big Data Analysis and Application, University of Science and Technology of China}
 }
 \email{qiliuql@ustc.edu.cn}

  \author{Fuzheng Zhang, Xing Xie}
 \affiliation{
 	\institution{Microsoft Research Asia}
 }
 \email{{fuzzhang,xing.xie}@microsoft.com}

 \author{Guoping Hu}
 \affiliation{
 	\institution{iFLYTEK Research}
 }
 \email{gphu@iflytek.com}

\renewcommand{\shortauthors}{Y. Yin et al.}

\begin{document}

\begin{abstract}
Transcribing content from structural images, e.g., writing notes from music scores, is a challenging task as not only the content objects should be recognized, but the internal structure should also be preserved. Existing image recognition methods mainly work on images with simple content (e.g., text lines with characters), but are not capable to identify ones with more complex content (e.g., structured code), which often follow a fine-grained grammar. To this end, in this paper, we propose a hierarchical \emph{S}potlight \emph{T}ranscribing \emph{N}etwork (STN) framework followed by a two-stage ``where-to-what'' solution. Specifically, we first decide ``where-to-look'' through a novel spotlight mechanism to focus on different areas of the original image following its structure. Then, we decide ``what-to-write'' by developing a GRU based network with the spotlight areas for transcribing the content accordingly. Moreover, we propose two implementations on the basis of STN, i.e., STNM and STNR, where the spotlight movement follows the Markov property and Recurrent modeling, respectively. We also design a reinforcement method to refine our STN framework by self-improving the spotlight mechanism. We conduct extensive experiments on many structural image datasets, where the results clearly demonstrate the effectiveness of STN framework.
\end{abstract}



\maketitle

\section{Introduction}
Transcribing content from images refers to recognizing semantic information in images into comprehensible forms (e.g., text) in computer vision~\cite{ye2015text}. It is an essential problem for computers to understand how humans communicate about what they see, which includes many tasks, such as reading text from scenes~\cite{zhang2013text,kannan2014mining}, writing notes from music scores~\cite{rebelo2012optical} and recognizing formulas from pictures~\cite{chan2000mathematical}. As it is crucial in many applications, e.g., image retrieval~\cite{cao2016deep,ShangLZYZW15}, online education systems~\cite{huang2017question,liu2018fuzzy} and assistant devices~\cite{ezaki2004text}, much attention has been attracted from both academia and industry~\cite{ye2015text}.

\begin{figure}
    \centering
      \hfill
    \subfigure[Music score example]{
        \includegraphics[scale=0.58]{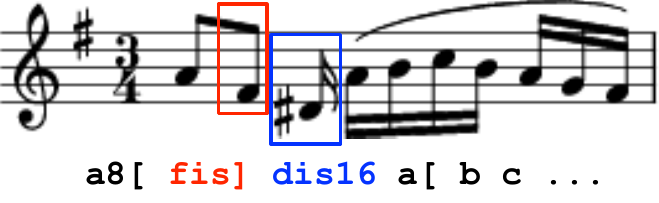}
        \label{fig:sub:eg1a}
    }
    \subfigure[Formula example]{
        \includegraphics[scale=0.58]{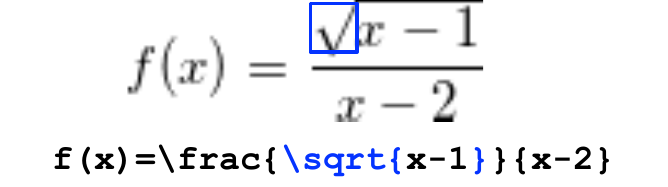}
        \label{fig:sub:eg1b}
    }
    \caption{Some structural image examples. Left is a music bar from Cello Suite No. 1 in G major by Bach; Right is a function formula from a high school math exercise.} \label{fig:eg}
\end{figure}

In the literature, there are many efforts for this transcribing problem, especially on text reading task. Among them, the most representative one called Optical Character Recognition (OCR) has been extensively studied in many decades~\cite{impedovo1991optical}, which mainly follows rule-based solutions for generating texts from well-scanned documents~\cite{lu2008document}. Recently, researchers focus on a more general scene text recognition task, aiming to recognize texts from natural images~\cite{vinyals2015show}. Usually, existing approaches are designed in an encoder-decoder architecture, which consists of two components: (1) a CNN based encoder to capture and represent images as feature vectors that preserve their the semantic information~\cite{oquab2014learning}; (2) a RNN based decoder that decodes the features and generates output text sequences either directly~\cite{vinyals2015show}, or attentively~\cite{xu2015show}. Though good performances have been achieved, previous studies mainly focus on the images with straightforward content (i.e., text with characters), while ignoring large proportion of structural images, where the content objects are well-formed in complex manners, e.g., music scores (Figure~\ref{fig:sub:eg1a}) and formulas (Figure~\ref{fig:sub:eg1b}). Therefore, the problem of transcribing content from these structural images remains pretty much open.

In fact, there are many technical challenges along this line due to the unique characteristics of structural images. First, different from natural images, where the text content is mostly placed in simple patterns, in structural images, the content objects usually follow a fine-grained grammar, and are organized in a more complex manner. E.g., in Figure~\ref{fig:sub:eg1a}, notes from the music score are not only placed simply from left to right, but the positions in the stave for each note are also specified, often with annotations added left or above. A division formula in Figure~\ref{fig:sub:eg1b} contains nested structure, where the equation components are placed at the left and right side of the equal sign, with two parts of the right-hand-side fraction placed above and below the middle line. Thus, it is necessary for transcribing to not only capture the information from local areas, but also preserve the internal structure and organization of the content. Second, content objects in structural images, even if they just take a small proportion, may carry much semantics. For example, the note marked by blue box in Figure~\ref{fig:sub:eg1a} is written as ``\texttt{dis16}'' in LilyPond\footnote{A domain specific language for music notation, http://lilypond.org/}, which means that the note is D\# (``-\texttt{is}'' for sharp), and the note is a sixteenth note (denoted by ``\texttt{16}''); the formula marked in Figure~\ref{fig:sub:eg1b} means ``\verb+\sqrt{...}+'' in \TeX\ code, representing the square root operator, with the scope defined by curly braces. Thus, it is very challenging to transcribe the complete content from an area containing such a informative object, compared to just one character in tasks such as scene text recognition. Third, there exist plenty of similar objects puzzling the transcribing task, e.g., a sixteenth note (blue in Figure~\ref{fig:sub:eg1a}) just contains one more flag on the stem than an eighth note (red), while notes with same duration and different pitches are almost identical except for their positioning. This characteristic requires a careful design for the transcribing.

To address the above challenges, following the observation on human transcribing process, i.e., first find out where to look, then write down the content, we present a two-stage ``where-to-what'' solution and propose a hierarchical framework called the \emph{S}potlighted \emph{T}ranscribing \emph{N}etwork (STN) for transcribing content from structural images. Specifically, after encoding images as features vectors, in our decoder component, we first propose a spotlight module with a novel mechanism to handle the ``where-to-look'' problem and decide a reading path focusing on areas of the original image following its internal structure. Then, based on the learned spotlights areas, we aim for ``what-to-write'' problem and develop a GRU based network for transcribing the semantic content from the local spotlight areas. Moreover, we propose two implementations on the basis of the STN framework. The first is a straightforward one, i.e., \emph{STNM with Markov property}, in which the spotlight placement follows a Markov chain. Comparatively, the second is a more sophisticated one, i.e., \emph{STNR with Recurrent modeling}, which can track long-term characteristics of spotlight movements. We also design a reinforcement method to refine STN, self-improving the spotlight mechanism. We conduct extensive experiments on real-world structural image datasets, where the results clearly demonstrate the effectiveness of the STN framework.


\section{Related Work}
The related research topics to our concerns can be classified into the following three categories: encoder-decoder system, attention mechanism, and reinforcement learning.

\subsection{Encoder-Decoder System} 
The encoder-decoder system is a general framework, which has been applied to many applications, such as neural machine translation~\cite{cho2014properties,bahdanau2014neural} and image captioning~\cite{vinyals2015show,xu2015show}. Generally, the system has two separate parts, one encoder for representing and encoding the input information into a feature vector, and one decoder for generating the output sequence according to the encoded representation. Due to its remarkable performance, many efforts have been made to apply it to scene text recognition~\cite{wang2012end}, aiming at transcribing texts from natural images. Specifically, for encoder design, representative works leveraged deep CNN based networks, which have been the most popular methods due to their performance on hierarchical feature extraction~\cite{oquab2014learning}, to learn the information encodings from images~\cite{jaderberg2016reading}. Then for decoder selection, variations of recurrent neural networks (RNN), such as LSTM~\cite{hochreiter1997long} and GRU~\cite{chung2014empirical}, were utilized to generate the output text sequence, both of which are able to preserve long-term dependencies for text representations~\cite{sundermeyer2012lstm}. The whole architecture is end-to-end, which show the effectiveness in practice~\cite{shi2017end}.

\subsection{Attention Mechanism} 
 However, in the original encoder-decoder systems, encoding the whole input into one vector usually makes the encoded information of images clumsy and confusing for the decoder to read from, leading to unsatisfactory transcription~\cite{luong2015effective}. To improve the encoder-decoder models addressing this problem, inspired by human visual system, researchers have tried to propose many attention mechanisms to highlight different parts of the encoder output by assigning weights to encoding vectors in each step of text generation~\cite{bahdanau2014neural,xu2015show,mnih2014recurrent} or sequential prediction~\cite{su2018exercise,ying2018sequential}. For example, Bahdanau et al.~\cite{bahdanau2014neural} proposed a way to jointly generate and align words using attention mechanism. Xu et al.~\cite{xu2015show} proposed soft and hard attention mechanisms for image captioning. Lee et al.~\cite{lee2016recursive} used an attention-based encoder-decoder system for character recognition problems.

Our work improves the previous studies mainly from the following two aspects. First, the attention weights are usually calculated by the correspondence between outputs and the whole content, which let the models know ``what'' to look but not ``where'' to look. In our work, we propose a novel spotlight mechanism to directly find a reading path tracking the image structure for transcribing. Second, previous decoding process has one RNN for learning attentions and transcribing simultaneously, which may cause some confusion for transcription, while our framework models spotlighting and transcribing with two separate facilities, avoiding the confusion between two sequences.

\begin{figure*}[ht]
	\centering
	\includegraphics[scale=0.54]{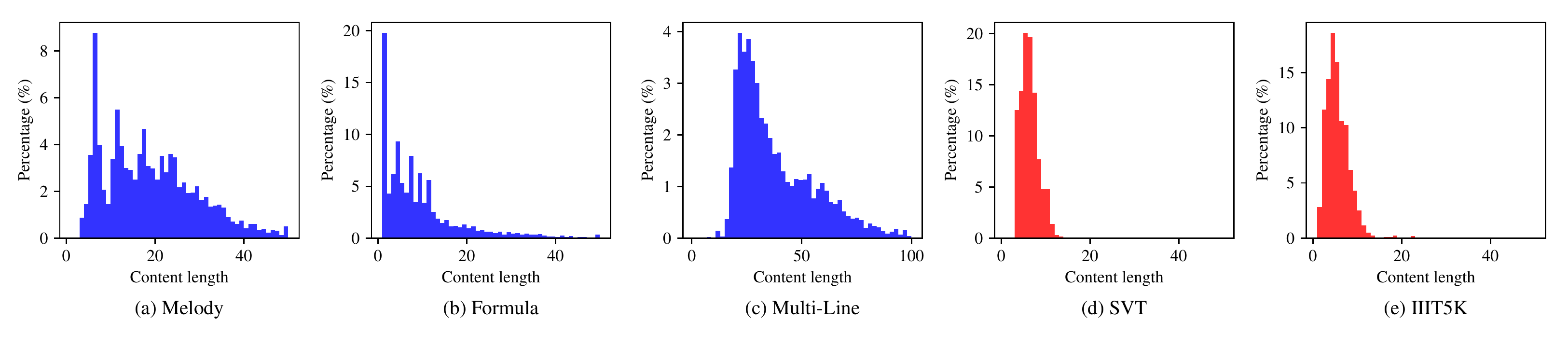}
	\caption{Comparison of structural image (blue) and scene text recognition datasets (red) on content length distribution.}\label{fig:dist}
\end{figure*}

\subsection{Reinforcement Learning}
Deep reinforcement learning is a kind of state-of-the-art technique, which has shown superior abilities in many fields, such as gaming and robotics~\cite{arulkumaran2017brief}. The main idea of them is to learn and refine model parameters according to task-specific reward signals. For example, Ranzato et al.~\cite{ranzato2015sequence} used the whole sequence metrics to guide the sequence generation, using REINFORCE method; Bahdanau et al.~\cite{bahdanau2016actor} utilized the actor-critic algorithm for sequence prediction, refining the model to improve sentence BLEU score.


\section{Preliminaries}
In this section, we first give a clear definition of structural images, and introduce the structural image datasets used in this paper. Then we discuss the crucial differences between structural image transcribing and typical scene text recognition with exclusive data analysis. At last, we give the formal definition of the structural image transcription problem.

\subsection{Data Description}
In this paper, we mainly focus on transcribing content from structural images. \textit{Structural images} refer to printed graphics that are not only a set of content objects, but also contain meaningful structure, i.e., object placement, following a certain grammar. Content with its structure can often be described by a domain specific language and complied by the corresponding software. Typical structural images include music scores, formulas and flow charts, etc., which can be described in music notation, \TeX\ and UML code, respectively.

\begin{table}[t] 
    \caption{The statistics of the datasets.}
    \label{tab:datastats}
    \centering
    \begin{tabular}{c|ccccc}
        \toprule
        Dataset & \thead{Image\\ count} & \thead{Token\\ space} & \thead{Token\\ count} & \thead{Avg. tokens \\ per image} & \thead{Avg. image\\ pixels} \\
        \midrule
        Melody & 4208 & 70 & 82,834 & 19.7 & 15,602.7 \\
        Formula & 61649 & 127 & 607,061 & 9.7 & 1,190.7 \\
        Multi-Line & 4595 & 127 & 182,112 & 39.8 & 9,016.6 \\
        \hline
        SVT & 618 & 26 & 3,796 & 5.9 & 12,733.5 \\
        IIIT5K & 3000 & 36 & 15,269 & 5.0 & 11,682.0 \\
        \bottomrule
    \end{tabular}
\end{table}

We exploit two real-world datasets, i.e., \textit{Melody} and \textit{Formula}, along with one synthetic dataset \textit{Multi-Line}, specifically for the structural image transcription task\footnote{Datasets are available at: http://home.ustc.edu.cn/\textasciitilde yxonic/stn\_dataset.7z.}. The \textit{Melody} dataset contains pieces of music scores and their source code in LilyPond collected from the Internet\footnote{http://web.mit.edu/music21/}, mostly instrumental solos and choral pieces written by Bach, split into 1 to 4 bar length, forming 4208 image-code pairs. The \textit{Formula} dataset is collected from Zhixue.com, an online educational system, which contains 61649 printed formulas from high school math exercises, with their corresponding \TeX\ code. To further demonstrate transcription on images with more complicated structure, we also construct the \textit{Multi-Line} dataset that contains 4595 multi-line formulas, e.g., piecewise function, each line consisting of some complex formulas, e.g., multiple integral. We summarize some basic statistics of these datasets in Table~\ref{tab:datastats}.

We now conduct deep analysis to show the unique characteristics of the structural image transcription task compared to traditional scene text recognition. Specifically, we compare our datasets with two commonly used datasets for scene text recognition, i.e., SVT~\cite{wang2011end} and IIIT5K~\cite{Mishra2012iiit5k}, and conclude three main differences. First, structural image transcription needs to preserve more information: other than just objects, how they are organized should also be transcribed. As shown in Table~\ref{tab:datastats} and Figure~\ref{fig:dist}, our datasets contain significantly longer content in relatively small images. Sequences longer than 10 tokens taking 75.0\%, 30.4\% and 99.9\% of Melody, Formula and Multi-Line datasets, respectively. However, only 1.9\% in SVT and 2.7\% in IIIT5K have more than 10 character long sequences. In addition, Melody, Formula and Multi-Line contain in average 1.26, 8.15 and 4.14 tokens every 1000 pixels, while SVT and IIIT5k only contain 0.46 and 0.43 characters, respectively, which indicates that each proportion of an image contains more information to be transcribed, along with the informative structure. Second, the output space and count in our datasets are often larger than SVT and IIIT5K, as shown in Table~\ref{tab:datastats}. Hence, it is even more complicated to transcribe content from structural images compared to text recognition. Third, structural image transcription process is reversible, meaning the corresponding code should be able to compile and regenerate the original image, which is not necessary or possible for traditional scene text recognition.

In summary, the above analysis clearly shows that the structural image transcription problem is quite different from traditional scene text recognition tasks. As a result, it is necessary to design a new approach that better fits this problem.

\begin{figure*}[ht]
    \centering
    \includegraphics[scale=0.65]{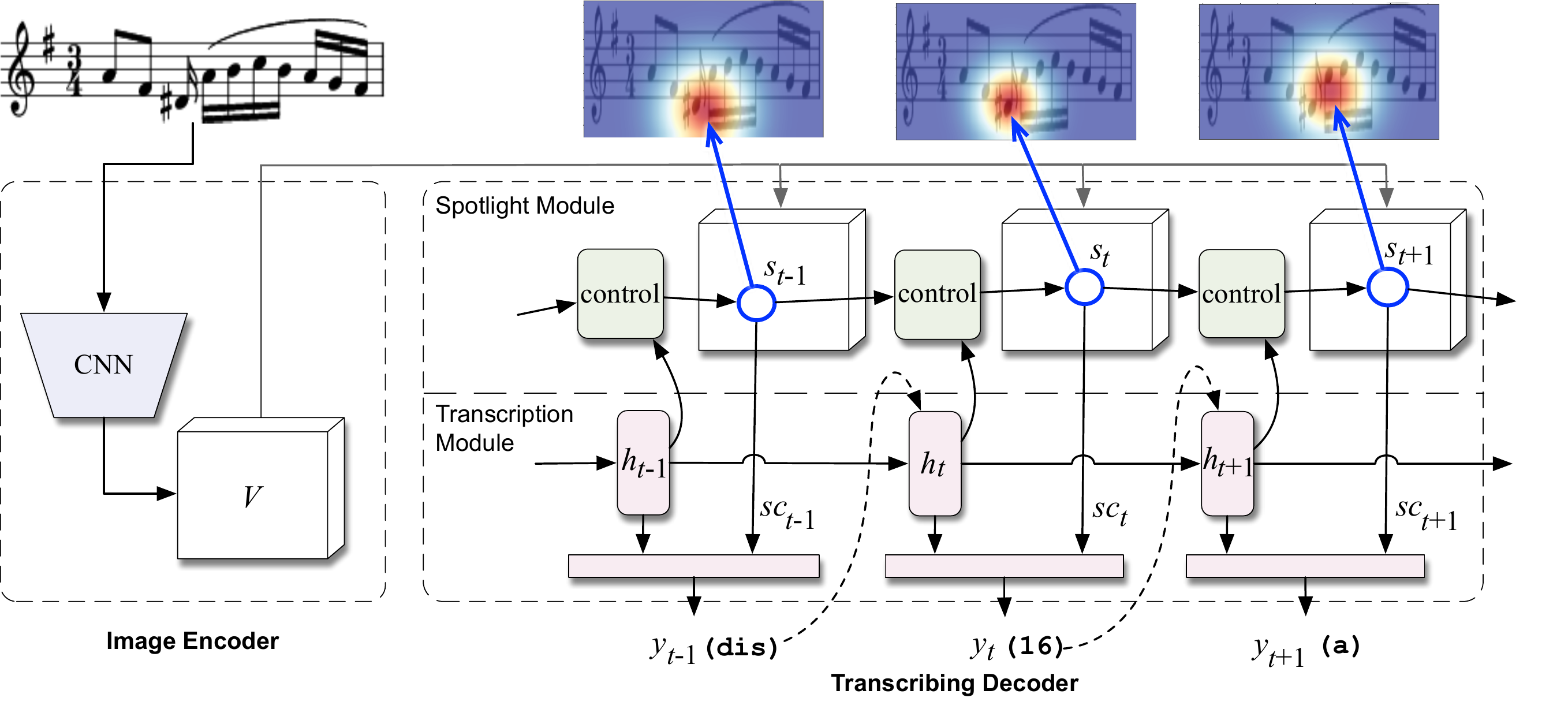}
    \caption{The STN model architecture consists of two main parts: 1) a convolutional image feature extractor as the encoder, and 2) the transcribing decoder. At the decoding stage, the spotlight module is first engaged to handle the ``where-to-look'' problem. Afterwards, the transcription module finds out ``what-to-write'' by utilizing the spotlighted information from the encoder, generating the transcribed content one token at a time. }
    \label{fig:model_arch}
\end{figure*}

\subsection{Problem Definition}
In this subsection, we formally introduce the structural image transcription problem. In our image transcribing applications, we are given structural images and their corresponding source code. Each input image $x$ is a one-channel gray-scale image with width $W$ and height $H$, containing content such as music notations or printed formulas. For each image, the expected output, i.e., its source code, is given as a token sequence $y=\{y_1, y_2, \ldots, y_T\}$, where $T$ is the length of token sequence. Each $y_t$ can be a LilyPond notation (\verb+c+, \verb+fis+, \ldots) in music score transcribing task, or a \TeX\ token (\texttt{x}, \verb+\frac+, \ldots) in formula transcribing task. Moreover, structural images are reversible, by which we mean that the token sequence is expected to reconstruct the original image using the corresponding compiler. Therefore, the problem can be defined as:

\begin{definition}({\small \textbf{Structural Image Transcription Problem}}).
Given a structural $W \times H$ image $x$, our goal is to transcribe the content from it as a sequence $\hat{y}=\{\hat{y}_1, \hat{y}_2, \ldots, \hat{y}_T\}$ as close as possible to the source code sequence $y$, where each $\hat{y}_t$ is the predicted token taking from the specific language corresponding to the image.
\end{definition}


\section{Spotlighted Transcribing Network}
In this section, we introduce the Spotlighted Transcribing Network (STN) framework in detail. First we give an overview of the model architecture. Then we describe all the details of our proposed spotlight mechanism in following sections. Finally we discuss the training process of STN with reinforcement learning for refinement.

\subsection{Model Overview}
Figure~\ref{fig:model_arch} shows the overall architecture of Spotlighted Transcribing Network (STN), which consists of two main components: (1) a convolutional feature extractor network as the encoder, which learns the visual representations $V$ from the input image $x$; (2) a hierarchical transcribing decoder, which we mainly focus on in this work. Mimicking human reading process, the decoder first takes the encoded image information $V$ and find out ``where-to-look'' by shedding spotlight on it, following the learned reading path, then generates the token sequence $y$, by predicting one token at a time using a GRU-based output network, solving the ``what-to-write'' problem. In the following subsections, we will explain how each part of the STN works in detail.

\subsection{Image Encoder}
The encoder part of STN is for extracting and embedding information from the image. Instead of embedding the complete image $x$ into one vector, which may cause a loss in structural information~\cite{xu2015show}, we extract a set of feature vectors $V$, each of which is a $D$-dimensional representation corresponding to a part of the image:
\[
V=\{V^{(i,j)}:i=1, \ldots, W',\,j=1,\ldots,H'\},\,V^{(i,j)}\in\mathbb{R}^D.
\]

A deep convolutional neural network (CNN) is used as the feature extractor to capture high-level semantic information, which we denote as $f(\cdot\,;\theta_f)$. We follow the state-of-the-art image feature extractor design as in ResNet~\cite{he2016deep}, adding residual connections between convolutional layers, together with ReLU activation~\cite{nair2010rectified} and batch normalization~\cite{ioffe2015batch} to stabilize training, but removing the fully connected layers along with higher convolutional and pooling layers. As a result, we construct an extractor network that takes an image $x$, outputs a 3 dimensional tensor $V$ ($W'\times H'\times D$):

\begin{equation}
V=f(x;\theta_f),
\end{equation}
where vector $V^{(i,j)}$ at each location $(i,j)$ represents the local semantic information. The output tensor also preserves spatial and contextual information, with the property that adjacent vectors representing neighboring parts of the image. This allows the decoder module to use the image information selectively with both content and location in mind.

\subsection{Transcribing Decoder}
The transcribing decoder of STN, as in typical encoder-decoder architecture, generates one token at a time, by giving its conditional probability over the encoder output $V$ and all the previous outputs $\{y_1,\ldots,y_{t-1}\}$ at each time step $t$. Hence, we can denote the probability of a decoder yielding a sequence $y$ as:
\begin{equation}
\mathrm{P}(y|x)=\prod_{t=1}^{T}{\mathrm{P}(y_t|y_1,\ldots,y_{t-1},V)}.
\end{equation}

Considering the fact that the output history can be long, we embed the history before time step $t$ into a hidden state vector $h_t$ by utilizing a variation of RNN --- Gated Recurrent Unit (GRU), which preserves more long-term dependencies. Formally, at time step $t$, the hidden state for output history $h_t$ is updated based on the last output item $y_{t-1}$ and the previous output history $h_{t-1}$, by an GRU network $GRU(\cdot\,;\theta_h)$:
\begin{equation}
h_{t}=GRU(y_{t-1}, h_{t-1};\theta_h).
\end{equation}

For image part, the visual representation $V$ we get as the encoder output carries enough semantic information, but as a whole it can be confounding for the decoder to comprehend, and thus needs careful selection~\cite{xu2015show}. To deal with this problem, we mimic what human do when reading images: focus on one spot at a time, write down content, then focus on a next spot following the image structure~\cite{blakemore1969existence}. Along this line, we propose a module with novel spotlight mechanism, where at each time step, we only focus on information around a certain spotlight center. We refer to the spotlight center position as $s_t$ at time step $t$, and the spotlighted information as spotlight context $sc_t$. Further details on how to get focused spotlight context are described in Section~\ref{spot}, while how to move the spotlight following the structure is described in Section~\ref{control}.

With embedded history $h_t$, and spotlight context $sc_t$, together with current spotlight position $s_t$, the conditional probability of output token at time $t$ can then be parameterized as follows:
\begin{equation}
\mathrm{P}(y_t|y_1,\ldots,y_{t-1}, V)=\mathrm{Softmax}(d(h_t\oplus sc_t\oplus
s_t;\theta_d)),
\end{equation}
where $d(\cdot\,;\theta_d)$ is a transformation function (e.g. a feed-forward neural network) that outputs a vocabulary-sized vector, and $\oplus$ represents the operation that concatenates two vectors. The overall transcription loss $\mathcal{L}$ on an image-sequence pair is then defined as the negative log likelihood of the token sequence over the image:
\begin{equation}\label{eq:loss}
\mathcal{L}=\sum_{t=1}^{T}{-\log{P(y_t|y_1,\ldots,y_{t-1},V)}}.
\end{equation}

With all the calculation being deterministic and differentiable, the model can be optimized through standard back-propagation. 

\begin{figure}
	\centering
	\begin{tikzpicture}[scale=0.67, every node/.style={scale=0.7}]
	\small
	\draw[step=0.5cm, thin] (0,0) grid (2,2);
	\path (-0.5, 1) node(par) {$\Bigg[\bigg($};
	\path (0.25,0.25) node(h) {1} 
	(0.25,1.75) node(1) {1}
	(0.25,1.25) node(2) {1}
	(0.25,0.75) node(d) {\ldots};
	\path (0.75,0.25) node(h) {2} 
	(0.75,1.75) node(1) {2}
	(0.75,1.25) node(2) {2}
	(0.75,0.75) node(d) {\ldots};
	\path (1.25,0.25) node(h) {\ldots} 
	(1.25,1.75) node(1) {\ldots}
	(1.25,1.25) node(2) {\ldots}
	(1.25,0.75) node(d) {\ldots};
	\path (1.75,0.25) node(h) {$W'$} 
	(1.75,1.75) node(1) {$W'$}
	(1.75,1.25) node(2) {$W'$}
	(1.75,0.75) node(d) {\ldots};
	\path (1,-0.25) node(x) {$I$};
	
	\path (2.25,1) node(x) {$-$};
	
	\draw[step=0.5cm, thin] (2.499,0) grid (4.5,2);
	\path (2.75,0.25) node(h) {$x_t$} 
	(2.75,1.75) node(1) {$x_t$}
	(2.75,1.25) node(2) {$x_t$}
	(2.75,0.75) node(d) {\ldots};
	\path (3.25,0.25) node(h) {$x_t$} 
	(3.25,1.75) node(1) {$x_t$}
	(3.25,1.25) node(2) {$x_t$}
	(3.25,0.75) node(d) {\ldots};
	\path (3.75,0.25) node(h) {\ldots} 
	(3.75,1.75) node(1) {\ldots}
	(3.75,1.25) node(2) {\ldots}
	(3.75,0.75) node(d) {\ldots};
	\path (4.25,0.25) node(h) {$x_t$} 
	(4.25,1.75) node(1) {$x_t$}
	(4.25,1.25) node(2) {$x_t$}
	(4.25,0.75) node(d) {\ldots};
	\path (3.5,-0.25) node(x) {$X_t$};
	\path (5, 1) node(par) {$\bigg)^2$};
	
	\path (5.35, 1) node(par) {$+$};
	
	\draw[step=0.5cm, thin] (0+5.99,0) grid (2+6,2);
	\path (-0.25+5.95, 1) node(par) {$\bigg($};
	\path (0.25+6,0.25) node(h) {$H'$} 
	(0.25+6,1.75) node(1) {1}
	(0.25+6,1.25) node(2) {2}
	(0.25+6,0.75) node(d) {\ldots};
	\path (0.75+6,0.25) node(h) {$H'$} 
	(0.75+6,1.75) node(1) {1}
	(0.75+6,1.25) node(2) {2}
	(0.75+6,0.75) node(d) {\ldots};
	\path (1.25+6,0.25) node(h) {\ldots} 
	(1.25+6,1.75) node(1) {\ldots}
	(1.25+6,1.25) node(2) {\ldots}
	(1.25+6,0.75) node(d) {\ldots};
	\path (1.75+6,0.25) node(h) {$H'$} 
	(1.75+6,1.75) node(1) {1}
	(1.75+6,1.25) node(2) {2}
	(1.75+6,0.75) node(d) {\ldots};
	\path (1+6,-0.25) node(x) {$J$};
	\path (2.25+6,1) node(x) {$-$};
	
	\draw[step=0.5cm, thin] (2.499+6,0) grid (4.5+6,2);
	\path (2.75+6,0.25) node(h) {$y_t$} 
	(2.75+6,1.75) node(1) {$y_t$}
	(2.75+6,1.25) node(2) {$y_t$}
	(2.75+6,0.75) node(d) {\ldots};
	\path (3.25+6,0.25) node(h) {$y_t$} 
	(3.25+6,1.75) node(1) {$y_t$}
	(3.25+6,1.25) node(2) {$y_t$}
	(3.25+6,0.75) node(d) {\ldots};
	\path (3.75+6,0.25) node(h) {\ldots} 
	(3.75+6,1.75) node(1) {\ldots}
	(3.75+6,1.25) node(2) {\ldots}
	(3.75+6,0.75) node(d) {\ldots};
	\path (4.25+6,0.25) node(h) {$y_t$} 
	(4.25+6,1.75) node(1) {$y_t$}
	(4.25+6,1.25) node(2) {$y_t$}
	(4.25+6,0.75) node(d) {\ldots};
	
	\path (3.5+6,-0.25) node(x) {$Y_t$};
	\path (5+6.35, 1) node(par) {$\bigg)^2\left.\Bigg]/\sigma_t^2\right.$};
	\end{tikzpicture}
	
	\caption{Demonstration of the parallelized operation on assigning weights. It should be clear that the element at each position $(i,j)$ of the result matrix is $[(i-x_t)^2+(j-y_t)^2]/\sigma_t^2$.} \label{fig:coord}
	
\end{figure}

\subsection{Spotlight Mechanism}\label{spot}
In this subsection, we describe how to get focused information of the input image, i.e., the spotlight context $sc_t$, with our proposed spotlight mechanism. How the spotlight moves through time is handled in a separate spotlight control module, and is described later in detail in Section~\ref{control}.

As mentioned earlier, the visual embedding $V$ is confounding for the decoder, and we want to focus on one spot at a time when generating output. To achieve this goal, we propose a novel spotlight mechanism to mimic human focus directly, where at each time step, we only care about information around a certain location which we call a spotlight center, by ``shedding'' a spotlight around it. More specifically, we define a spotlight handle $s_t=(x_t, y_t, \sigma_t)^\text{T}$ at each time step $t$ to represent the spotlight, where $(x_t,y_t)$ represents the center position of the spotlight, and $\sigma_t$ represents the radius of the spotlight. Inspired by Yang et al.~\cite{yang2017learning}, we ``shed'' a spotlight by assigning weights to image representation vectors at each position, following a truncated Gaussian distribution centered at $(x_t, y_t)$, with the same variance $\sigma_t$ on both axis.

Formally, under the spotlight with handle $s_t=(x_t, y_t, \sigma_t)^\text{T}$, the weights for each vector at position $(i, j)$ at time step $t$, denoted as $\alpha_t^{(i,j)}$, is proportional to the probability density at point $(i, j)$ under Gaussian distribution:
\begin{equation}
\alpha_t^{(i,j)} \sim \mathcal{N}((i,j)^\text{T}|\mu_t, \Sigma_t),
\end{equation}
\begin{equation}
\mu_t=(x_t, y_t)^T\quad \Sigma_t=\begin{bmatrix}
    \sigma_t & 0 \\
    0 & \sigma_t 
\end{bmatrix}.
\end{equation}

Intuitively, the closer $(i,j)$ is to the center $(x_t,y_t)$, the higher the weight should be, mimicking shedding a spotlight with radius $\sigma_t$ onto the location $(x_t,y_t)$. To calculate the weight $\alpha_t^{(i,j)}$ of each position $(i,j)$ while still make the process differentiable, we apply the definition of Gaussian distribution and rewrite the expression of $\alpha_t^{(i,j)}$ as:
\begin{equation}
\alpha_{t}^{(i, j)}=\mathrm{Softmax}(b_t)=\frac{\exp(b_t^{(i,j)})}{\sum_{u=1}^{W'}\sum_{v=1}^{H'}
{\exp(b_t^{(u,v)})}},
\end{equation}
\begin{equation}
b_t^{(i,j)}=-\frac{(i-x_t)^2+(j-y_t)^2}{\sigma_t^2}
\label{eq:a},
\end{equation}
where $b$ measures how close the point $(i,j)$ is to the center $(x_t,y_t)$, i.e., how important this point is, and $\alpha$ is thus a $W'\times H'$ matrix following the truncated Gaussian distribution for each point $(i,j)$, and can later be used as weights for each image feature vector.

To parallize the calculation of Equation (\ref{eq:a}), we perform a small trick as demonstrated in Figure~\ref{fig:coord}. We first construct two $W'\times H'$ matrices $I$ and $J$ in advance, each of them representing one coordinate. Specifically, as shown in Figure~\ref{fig:coord}, for each point $(i,j)$, we have $I^{(i,j)}=i$ and $J^{(i,j)}=j$. We also expand $x_t$ and $y_t$ as $W'\times H'$ matrices $X_t$ and $Y_t$ respectively, with same value for each element. Therefore, Equation~(\ref{eq:a}) can be written as the matrix form:
\begin{equation}
b_t=-[(I-X_t)^2+(J-Y_t)^2]/\sigma_t^2
\end{equation}

The focused information of the visual representation $V$ at time step $t$ can then be computed as a spotlight context vector $sc_t$ weighted by $\alpha_t^{(i,j)}$ according to current spotlight handle $s_t$, i.e., the weighted sum of features at each position:
\begin{equation}
sc_t = \sum_{i=1}^{W'}\sum_{j=1}^{H'}{ \alpha_t^{(i,j)} V^{(i,j)}}
\end{equation}

Please note that the spotlight context $sc_t$ represents the information in the focused area at time step $t$, and should contain useful information specifically for transcribing at current time step. By focusing directly on the correct spot, the transcription module therefore only cares about the local information, not confusing at areas with similar content all over the image.

\subsection{Spotlight Control}\label{control}
Now we discuss how to control the spotlight to find a proper reading path, following the image structure through the whole generation process. Different from traditional attention strategy where both output sequence and attention behavior are embedded in one module, we see the spotlight movement (i.e., the value of the spotlight handle $s_t=(x_t, y_t, \sigma_t)^\text{T}$ at each time step $t$) as a separate sequence devoted to following the image structure, and model this sequence with a standalone spotlight controlling module, without mixing the information with the output sequence. We provide two implementations under the STN framework, i.e., the straightforward \textit{STNM with Markov property}, and the more sophisticated \textit{STNR with Recurrent modeling}, utilizing another GRU network. Each implementation models the spotlight handle sequences differently.

\begin{figure}
	\centering
	\includegraphics[scale=0.6]{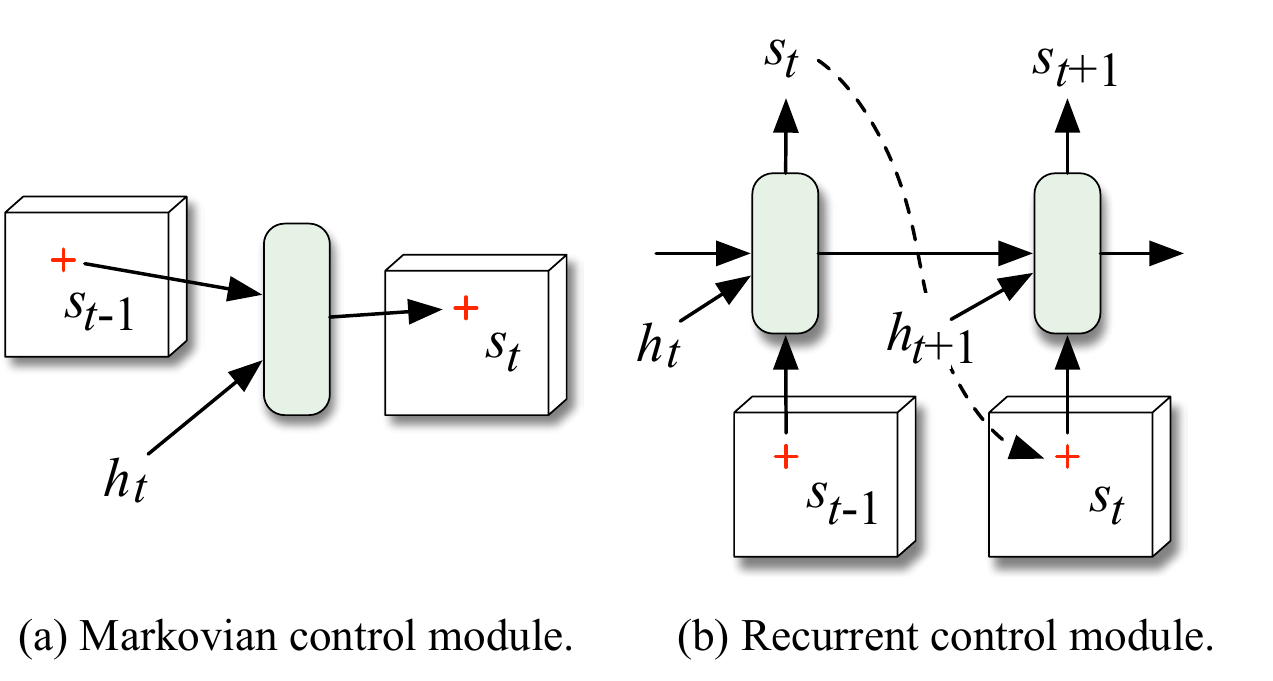}
	\caption{The spotlight control module implementations.}\label{fig:control}
\end{figure}

\textbf{STNM with Markov property.} With an assumption that is not far from reality, we can intuitively treat the spotlight handle sequence as a Markov process, i.e., current spotlight handle only depends on the previous handle, along with other internal states at current time step. Treating the spotlight handle as a Markov process means the probability of choosing $s_t$ at time $t$ does not rely on spotlight handles more than one step earlier, i.e.:
\begin{equation}
P(s_t|s_1,\ldots,s_{t-1};\cdot)=\\P(s_t|s_{t-1};\cdot).
\end{equation}
To decide where to put the spotlight properly, the model also needs to know current internal states at time step $t$, including the spotlight context $sc_{t-1}$ which represents previous spotlighted region, and the history embedding $h_t$ which represents output history \emph{before} time $t$. Thus, we can use a feed-forward neural network $n(\cdot\,;\theta_{n})$ to model the choice of $s_t$ (Figure~\ref{fig:control} (a)) as:
\begin{equation}
s_t=n(s_{t-1}\oplus sc_{t-1}\oplus h_{t}; \theta_{n})
\end{equation}
The way we model the sequence is simple and time-independent, which makes it easier for the controlling module to train.

\textbf{STNR with Recurrent modeling.} Sometimes longer spotlight history is needed for spotlight controlling on images with more complex structure. To track the image structure as a sequence with long-term dependency, we propose another GRU network $GRU(\cdot\,;\theta_g)$ to track the spotlight history, and a fully connected layer $c(\cdot\,;\theta_c)$ to generate next spotlight handle (Figure~\ref{fig:control} (b)). Specifically, at time step $t$, with last spotlight history embedding denoted as $e_{t}$, the current spotlight handle $s_t$ at time $t$ is calculated as:
\begin{equation}
s_t = c(e_{t}\oplus sc_{t-1}\oplus h_{t}; \theta_c)
\end{equation}
and the history embedding is updated by:
\begin{equation}
e_t = GRU(s_{t-1}, e_{t-1};\theta_g)
\end{equation}

Through a separate module specifically for spotlight control, STN gains two advantages over the traditional attention mechanism. First, STN focuses on local areas by design, and the model will only have to learn where to focus and what to transcribe, while the attention model have to first learn to focus, then learn what to focus on. Second, modeling reading and writing process as two separate sequences, with a standalone module dedicated for the ``where-to-look'' problem, STN is capable for directly learning a reading path on structural images apart from generating the output sequences, which enables our model to track the image structure more closely compared to attentive models where attentions and transcribing process are modeled together in only one network.

\subsection{Training and Refining STN}\label{training}

Parameters to be updated in both implementations comes from three parts: the encoder parameters $\theta_f$, the decoder parameters $\{\theta_h, \theta_d\}$, and parameters in the spotlight control module, which are $\theta_n$ in STNM and $\{\theta_c, \theta_g\}$ in STNR. The parameters are updated to minimize the total transcription loss $\mathcal{L}$ (Equation~(\ref{eq:loss})) through a gradient descent algorithm, which we choose the Adam optimizer~\cite{kingma2014adam}. More detailed settings are presented in the experiment section.

Though our model is differentiable, and can be optimized through back-propagation methods, directly training to fit the label suffers from some specific aspects in the image transcribing task. Firstly, the model has to jointly learn two different sequences with only one of them directly supervised, which may result in inaccurate reading path. Second, the given token sequence may only be one of the many correct ones that all regenerates the original image. For instance, in LilyPond notation, we can optionally omit duration for notes at same length with their predecessors. Fitting to only one of the correct sequences lets down the model even when it achieves good strategies. Fortunately, in structural image transcription problems, we have an advantage that the process is reversible, meaning given the transcribed sequence, we can use a compiler to reconstruct the image. With the guidance of this, we can further refine our model using reinforcement learning, by regarding our sequential generation as a decision making problem, viewing it as a Markov Decision Process (MDP)~\cite{bahdanau2016actor}. Formally, we define the \textit{state}, \textit{action} and \textit{reward} of the MDP as follows:

\textbf{State:} View our problem as outputting the probability of items at each time step conditioned by the image and previous generations, the environment state at time step $t$ as the combination of the image $x$ and the output history $\{y_1, \ldots, y_{t-1}\}$, which is exactly the inputs of the STN. Therefore, instead of directly using the environment state, we use the internal states (combined and denoted as $state_t$) in STN framework as MDP states.

\textbf{Action:} Taking action $a_t$ is defined as generating the token $y_t$ at time step $t$. With the probability of each token as the output, the STN can be viewed as a stochastic policy that generates actions by sampling from the distribution $\pi(a|state_t;\theta)=P(a|y_1, \ldots, y_{t-1}, x;\theta)$, where $\theta$ is the set of model parameters to be refined.

\textbf{Reward:} After taking the action, a reward signal $r$ is received. Here we define the reward $r_t$ as 0 when the generation is not finished at time step $t$, or the pixel similarity between the reconstruction image and the original image after the whole generation process finished. Besides, we give -1 as the final reward if the output sequence does not compile, addressing grammar constraints by penalizing illegal outputs. The goal is to maximize the sum of the discounted rewards from each time $t$, i.e., the return:
\begin{equation}
R_t=\sum_{k=t}^T{\gamma^kr_k}.
\end{equation}

We further define a value network $v(\cdot\,;\theta_v)$ for estimation of the expected return from each $state_t$, which is a feed-forward network with the same input as the STN output layer $d$. The estimated value $v_t$, i.e., the expected return, at time step $t$ is then
\begin{equation}
v_t=v(h_t\oplus sc_t\oplus s_t;\theta_v).
\end{equation}
With a stochastic policy together with a value network, we can apply the actor-critic algorithm~\cite{bahdanau2016actor} to our sequence generation problem, with the policy network trained using policy gradient at each time step $t$ as:
\begin{equation}
\nabla_\theta=\log\pi(a|state_t;\theta)(R_t-v_t),
\end{equation}
and the value network trained by optimizing the distance between the estimated value and actual return: $\mathcal{L}_{value}=||v_t-R_t||_2^2$.

As the whole model is complicated, directly applying reinforcement learning to the model suffers from the large searching space. Through experiments we notice that, after supervised training, the image extractor and the output history embedding modules have both been trained properly, and it is more important for our framework to have a better reading path to make precise predictions, which indicates that refining the spotlight module is most beneficial. Therefore, at reinforcement stage, we only optimize parameters from the spotlight control module ($\theta_n$ in STNM, $\theta_c$ and $\theta_g$ in STNR), along with those from the output layer ($\theta_o$), and omit $\theta_f$ and $\theta_h$, which reduces the variance when applying reinforcement learning algorithms, and get better improvements.

With this train-and-refine procedure, our model can learn a reasonable reading path on structural images, focusing on different parts following the image structure when transcribing, and get superior transcription results, as our experimental results show in the next section.

\section{Experiments}
In this section, we conduct extensive experiments to demonstrate the effectiveness of STN model from various aspects: (1) the transcribing performance; (2) the validation loss demonstrating the model sensitivity; (3) the spotlight visualization of STN.

\begin{table*} 
    \caption{Transcription accuracy on three datasets.} \label{tab:acc} 
    \centering
    \subtable[\textbf{Melody}]{
        \begin{tabular}{p{1.1cm}|p{0.7cm}|p{0.7cm}|p{0.7cm}|p{0.7cm}}
            \toprule
            \multirow{2}{*}{Baseline} &\multicolumn{4}{c}{Testing set percentage}\\
            \cline{2-5}& 40\%&30\%&20\%&10\%\\
            \midrule
            EncDec & 0.266&0.272&0.277&0.282\\
            AttnDot &0.524&0.548&0.580&0.617 \\
            AttnFC & 0.683&0.710&0.730&0.756 \\
            AttnPos & 0.725&0.736&0.741&0.758 \\
            STNM & 0.729&0.733&0.749&0.759 \\
            STNR & \textbf{0.738}&\textbf{0.748}&\textbf{0.758}&\textbf{0.767} \\
            \bottomrule
        \end{tabular}
        \label{tab:melody_acc}  
    }  
    \subtable[\textbf{Formula}]{  
        \begin{tabular}{p{1.1cm}|p{0.7cm}|p{0.7cm}|p{0.7cm}|p{0.7cm}}
            \toprule
            \multirow{2}{*}{Baseline} &\multicolumn{4}{c}{Testing set percentage}\\
            \cline{2-5}&
            40\%&30\%&20\%&10\%\\ \midrule
            EncDec & 0.405&0.427&0.445&0.451\\
            AttnDot & 0.530&0.563&0.600&0.611 \\
            AttnFC & 0.657&0.701&0.717&0.725\\
            AttnPos & 0.716&0.723&0.732&0.741 \\
            STNM & 0.717&0.726&0.740&0.749 \\
            STNR & \textbf{0.739}&\textbf{0.751}&\textbf{0.759}&\textbf{0.778} \\
            \bottomrule
        \end{tabular}
        \label{tab:formula_acc} 
    }
    \subtable[\textbf{Multi-Line}]{
        \begin{tabular}{p{1.1cm}|p{0.7cm}|p{0.7cm}|p{0.7cm}|p{0.7cm}}
            \toprule
            \multirow{2}{*}{Baseline} &\multicolumn{4}{c}{Testing set percentage}\\
            \cline{2-5}&
            40\%&30\%&20\%&10\%\\ \midrule
            EncDec & 0.218&0.227&0.251&0.267\\
            AttnDot & 0.334&0.447&0.554&0.599 \\
            AttnFC & 0.614&0.642&0.686&0.707\\
            AttnPos & 0.624&0.652&0.698&0.720 \\
            STNM & 0.674&0.705&0.731&0.734 \\
            STNR & \textbf{0.712}&\textbf{0.736}&\textbf{0.754}&\textbf{0.760} \\
            \bottomrule
        \end{tabular}
        \label{tab:multiline_acc} 
    }
\end{table*}

\begin{figure*}[ht]
    \centering
    \subfigure[Melody]{
    \includegraphics[scale=0.55]{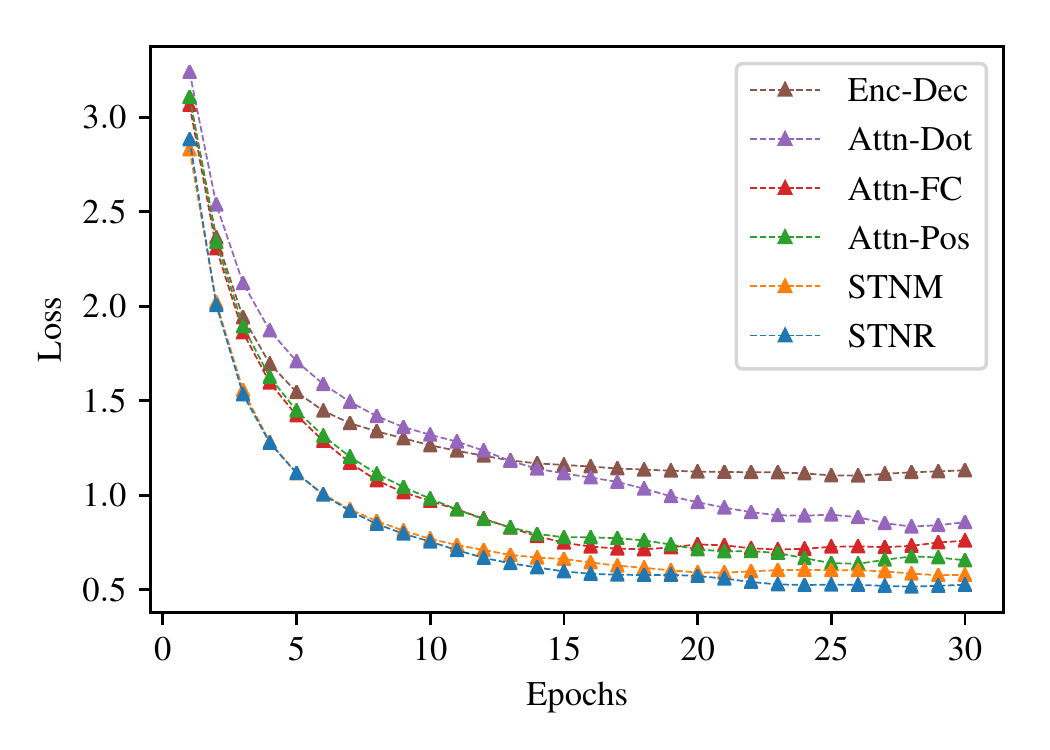}
    }
    \hspace{-1em}
    \subfigure[Formula]{
    \includegraphics[scale=0.55]{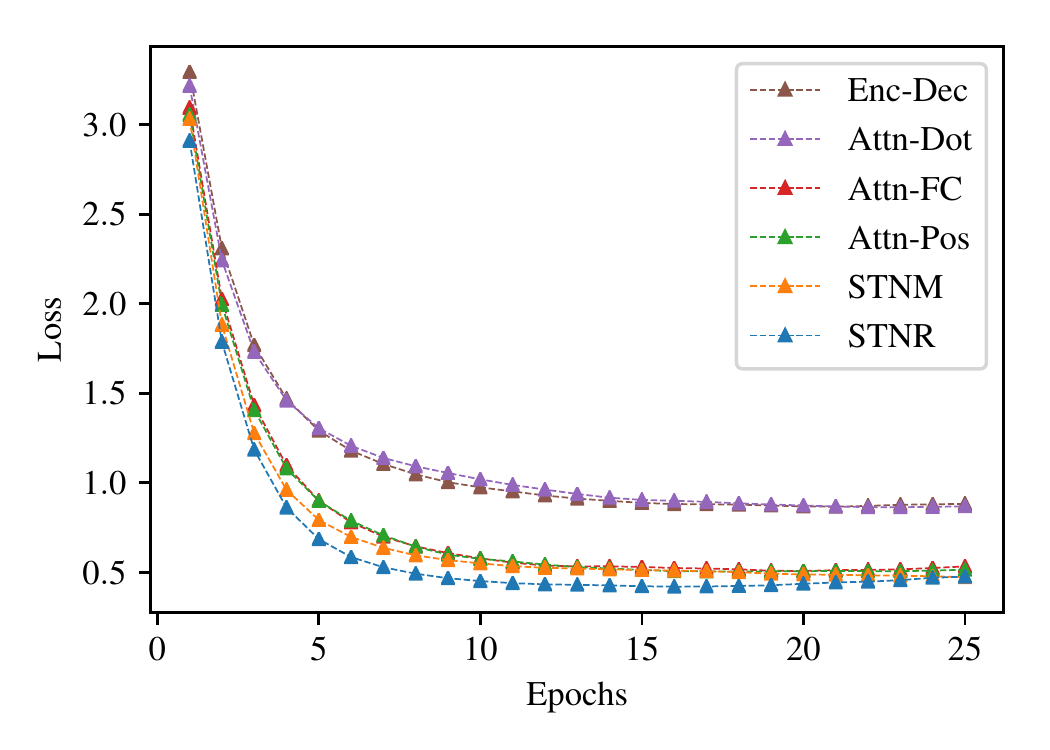}
    }
    \hspace{-1em}
    \subfigure[Multi-Line]{
    \includegraphics[scale=0.55]{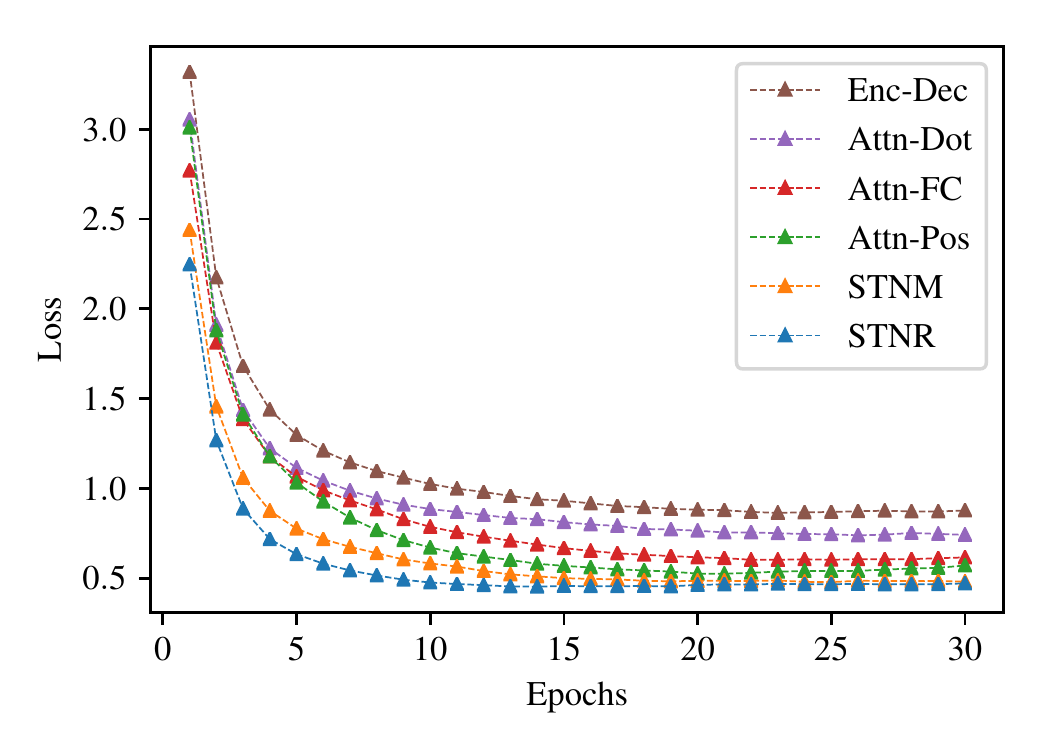}
    }
    
    \caption{Validation loss of all models on three datasets.}\label{fig:loss}

\end{figure*}

\subsection{Experimental Setup}
\subsubsection{Data partition and preprocessing.} We partition all our datasets, i.e., \textit{Melody}, \textit{Formula} and \textit{Multi-Line}, into 60\%/40\%, 70\%/30\%, 80\%/20\%, 90\%/10\% as training/testing sets, respectively, to test model performance at different data sparsity. From each training set, we also sample 10\% images as validation set. The images are randomly scaled and cropped for stable training, and ground-truth source code is cut into token sequences in the corresponding language to reduce searching space.

\subsubsection{STN setting.} 
We now specify the model setup in STN, including image encoder, transcription decoder and reinforcement module. For STN image encoder, we use a variation of ResNet~\cite{he2016deep}, and set the encoded vector width as 128. For its transcribing decoder, we set the output history embedding $h_t$, and the spotlight history embedding $e_t$ as the same dimensions of 128, respectively. The value network used at the reinforcement stage is a two-layer fully-connected neural network, with the hidden layer also sized at 128.

\subsubsection{Training setting.} 
To set up the training process, we initialize all parameters in STN following~\cite{glorot2010understanding}. Each parameter is sampled from $U\left(-\sqrt{6/(n_{in}+n_{out})},\sqrt{6/(n_{in}+n_{out})}\right)$ as their initial values, where $n_{in}$, $n_{out}$ stands for the number of neurons feeding in and neurons the result is fed to, respectively. Besides, to prevent overfitting, we also add L2-regularization term in the loss function (Equation~(\ref{eq:loss})), with the regularization amount adjusted to the best performance. At reinforcement stage, the discount factor $\gamma$ is set as 0.99. We also apply some techniques mostly mentioned in~\cite{bahdanau2016actor} to reduce variance, including using an additional target Q-network and reward normalization.

\subsubsection{Comparison methods.} 
To demonstrate the effectiveness of STN, we compare our two implementations, i.e., STNM and STNR, with many state-of-the-art baselines as follows.

\begin{itemize}
    \setlength{\itemsep}{0pt}
    \setlength{\parsep}{0pt}
    \setlength{\parskip}{0pt}
    \item \textbf{Enc-Dec} is a plain encoder-decoder model used originally for image captioning~\cite{vinyals2015show}. Its design allows it to be used in our problem setup with minor adjustments.
    \item \textbf{Attn-Dot} is an encoder-decoder model with attention mechanism following~\cite{luong2015effective}, where the attention score is calculated by directly computing the similarity between current output state and each encoded image vectors.
    \item \textbf{Attn-FC} is an encoder-decoder model similar to~\cite{vinyals2015show}, but with basic visual attention strategy. The model presents two attention strategies, i.e., the ``hard'' and ``soft'' attention mechanism, from which we follow~\cite{xu2015show} and choose the more widely used ``soft'' attention as it is deterministic and easier to train.
    \item \textbf{Attn-Pos} is an encoder-decoder model designed specifically for scene text recognition~\cite{yang2017learning}, where besides the image content, it also embeds location information into attention calculation, and get superior results.
\end{itemize}

To conduct a fair comparison, the image encoders for baselines are changed to use the more recent ResNet~\cite{he2016deep} as our model does, with all of them tuned to have the best performance. All models are implemented by PyTorch\footnote{http://pytorch.org}, and trained on a Linux server with four 2.0GHz Intel Xeon E5-2620 CPUs and a Tesla K20m GPU.

\subsection{Experimental Results}
\subsubsection{Transcribing performance}
We train STN along with all the baseline models on four different data partition of each, comparing token accuracy at different data sparsity. We repeat all experiments 5 times and report the average results which are shown in Table~\ref{tab:acc}.

From the results, we can get several observations. First, both STNM and STNR perform better than all the other methods. This indicates that STN framework is more capable for structural image transcription tasks, being more effective and accurate on tracking complex image structures. Second, STN models, as well as attention based methods, all have much higher prediction accuracy than plain EncDec method, which proves the claim mentioned earlier in this paper that image information encoded as a single vector is confounding for decoder to decode, and both STN and attentive models are able to reduce the confusion. Moreover, STN models are consistently better than those attentive ones, showing the superiority of STN with separate modules for spotlighting and transcribing. Third, STNR and STNM has slightly higher performance on \textit{Melody} and \textit{Formula} as Attn-Pos, but surpasses it marginally on \textit{Multi-Line} dataset. These results demonstrate that STN with spotlight mechanism can well preserve the internal structure of images, especially in more complex scenarios, benefiting the transcription accuracy. Last but not least, we can see that STNR consistently outperforms than STNM, which indicates that it is effective to track long-term dependency for spotlighting in the process of transcribing structural image content.

\begin{figure*}
	\centering
	\includegraphics[scale=0.44]{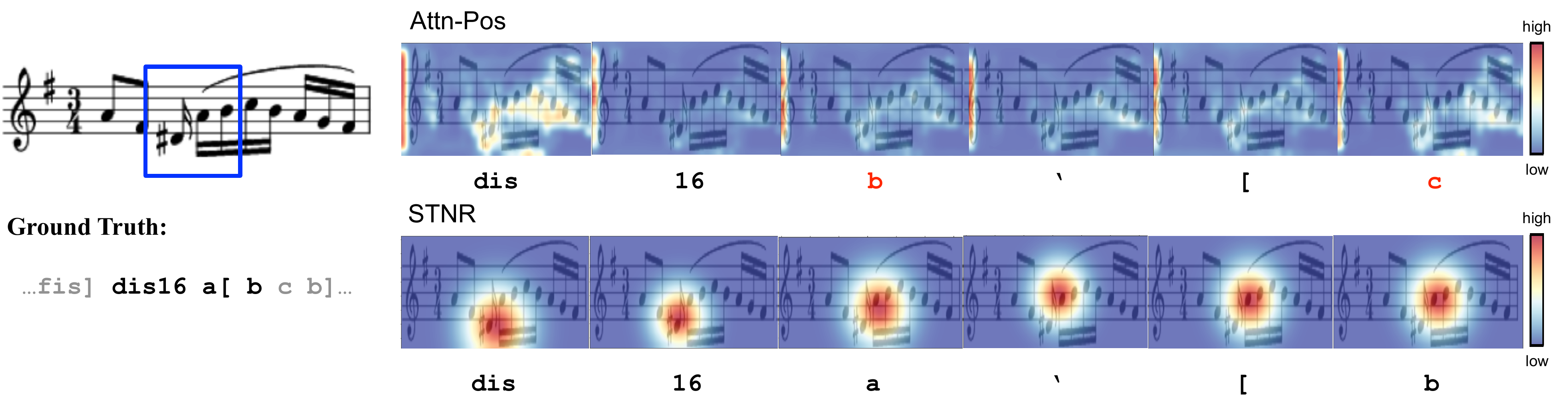}
	\caption{Comparison between attention and spotlight mechanism on Melody dataset.} \label{fig:vis_melody}
\end{figure*}
\begin{figure*}[ht]
    \centering
    \includegraphics[scale=0.44]{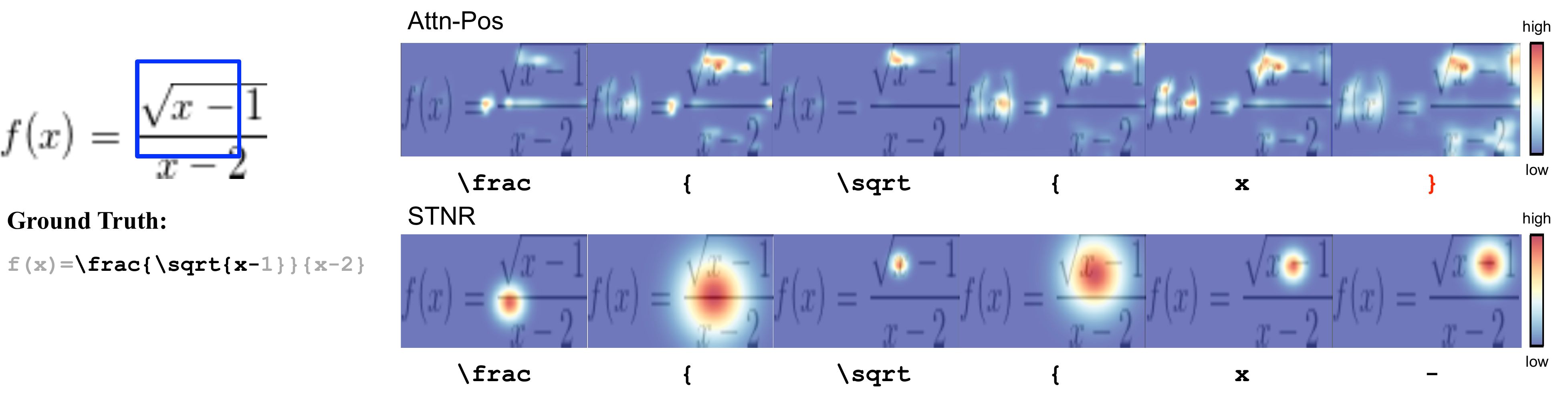}
    \caption{Comparison between attention and spotlight mechanism on Formula dataset.} \label{fig:vis_formula}
\end{figure*}

\subsubsection{Validation loss}
The losses of all models on the validation set throughout the training process on three datasets are shown in Figure~\ref{fig:loss}. There are also similar observations as before, which demonstrates the effectiveness of STN framework again. Clearly, from the results, both STNR and STNM converge faster than the other models, and also achieve a lower loss. Especially, the improvements of them on the more complex Multi-Line datasets are more significant. Thus, we can reach a conclusion that STN with spotlight mechanism has superior ability to transcribe content from structural images. Moreover, all models reach their lowest validation loss before 30 epochs, with STNR and STNM both come to their best point earlier. Thus, in our experiments, we train both STNR and STNM for 25, 15, 20 epochs on Melody, Formula and Multi-Line datasets respectively to obtain the best performance.

\subsubsection{Spotlight visualization}
To show the effectiveness of STN capturing the image structure and producing a reasonable reading path while transcribing, we visualize the spotlight weights computed by STNR when generating tokens, and compare them with the attention weights calculated by Attn-Pos model.

Figure~\ref{fig:vis_melody} and Figure~\ref{fig:vis_formula} visualize the results throughout image examples from Melody and Formula datasets, respectively.\footnote{We only choose two real-world datasets for visualization due to the page limitation.} In each example, we compare the attention and spotlight mechanism on how focused they are when generating a token, also on how well they track the image structure. From the visualization, we can draw conclusions that: (1) STNR finds a more reasonable reading path on both examples. In the melody example, it focuses on notes from left to right, and also tracks the height of each note, making accurate note pitch prediction; In the formula example, it clearly follows middle-top-bottom order when reading a fraction. Attn-Pos model on the other hand, does not track the image structure well enough. As shown in Figure~\ref{fig:vis_formula}, it fails to find the correct spot after generating ``\verb+\sqrt{x+'', losing track of the radical expression, and generates the wrong token ``\texttt{\}}'' at last. (2) Although Attn-Pos model assigns more weights on content objects in images, e.g., notes, formulas and variables, it is often confused at areas with similar content. On the other hand, STNR clearly distinguishes similar regions properly. More specifically, in Figure~\ref{fig:vis_melody}, although Attn-Pos is able to focus on the notes, all notes are given similar weights as they look similar, which causes confusion and then wrong prediction. And in Figure~\ref{fig:vis_formula}, when Attn-Pos writes \verb+x+, three \verb+x+'s in the image all have high weights, causing the model to forget where to look next. On the contrary, STNR is well focused on the correct spot when generating each token on both of the datasets, which leads to more precise predictions.

\subsubsection{Discussion}
All the above experiments have shown the effectiveness of STN on structural image transcription tasks. It has superior performance on structural image transcription task compared to other general-purpose approaches, and also captures the structure of the image by producing a reading path following the image structure when transcribing.

There are still some directions for further studies. First, STN learns to transcribe tokens directly with little prior knowledge of the image or specific languages. We are willing to utilize more prior knowledge, such as lexicons and hand-engineered features, to further improve the performance. Second, we will try to apply our model to some more ambitious settings, such as transcribing with long-term context, also to make our model capable for other transcribing applications such as scene text recognition. Third, we would like to further decouple the reading and writing process of STN, in order to mimic human behavior more genuinely.


\section{Conclusion}

In this paper, we presented a novel hierarchical Spotlighted Transcribing Network (STN) for transcribing content from structural images by finding a reading path tracking the image internal structure. Specifically, we first designed a two-stage ``where-to-what'' solution with a novel spotlight mechanism dedicated for the ``where-to-look'' problem, providing two implementations under the framework, modeling the spotlight movement through Markov chain and recurrent dependency, respectively. Then, we applied supervised learning and reinforcement learning methods to accurately train and refine the spotlight modeling, in order to learn a reasonable reading path. Finally, we conducted extensive experiments on one synthetic and two real-world datasets to demonstrate the effectiveness of STN framework with fast model convergence and high performance, and also visualized the learned reading path. We hope this work could lead to more studies in the future.

\section*{Acknowledgements}

This research was partially supported by grants from the National Natural Science Foundation of China (No.s U1605251 and 61727809), the Science Foundation of Ministry of Education of China \& China Mobile (No. MCM20170507), and the Youth Innovation Promotion Association of Chinese Academy of Sciences (No. 2014299).

\bibliographystyle{ACM-Reference-Format}
\bibliography{kdd}

\end{document}